\documentclass[conference]{IEEEtran}
\IEEEoverridecommandlockouts
\usepackage{booktabs}
\usepackage{algorithm}
\usepackage{algpseudocode}
\usepackage{cite}
\usepackage{amsmath,amssymb,amsfonts}
\usepackage{tabularx}
\usepackage{graphicx}
\usepackage{textcomp}
\usepackage{xcolor}
\usepackage{float}
\usepackage{listings}
\usepackage{balance}
\usepackage[font=footnotesize,labelfont=bf]{caption}
\DeclareCaptionLabelFormat{IEEE}{#1~#2}
\def\BibTeX{{\rm B\kern-.05em{\sc i\kern-.025em b}\kern-.08em
    T\kern-.1667em\lower.7ex\hbox{E}\kern-.125emX}}
\begin{document}

\title{A Two-Stage, Object-Centric Deep Learning Framework for Robust Exam Cheating Detection\thanks{Accepted at the FISU Joint Conference on Artificial Intelligence (FJCAI 2026), Vietnam.}
}

\author{\IEEEauthorblockN{Van-Truong Le}
\IEEEauthorblockA{\textit{University of Science, VNUHCM}\\
\textit{Vietnam National University} \\
Ho Chi Minh, Vietnam \\
23120181@student.hcmus.edu.vn}
\and
\IEEEauthorblockN{Le-Khanh Nguyen}
\IEEEauthorblockA{\textit{University of Science, VNUHCM}\\
\textit{Vietnam National University} \\
Ho Chi Minh, Vietnam \\
23120052@student.hcmus.edu.vn}
\and
\IEEEauthorblockN{Trong-Doanh Nguyen}
\IEEEauthorblockA{\textit{University of Science, VNUHCM}\\
\textit{Vietnam National University} \\
Ho Chi Minh, Vietnam \\
23120004@student.hcmus.edu.vn}
}

\maketitle

\begin{abstract}
Academic integrity continues to face the persistent challenge of examination cheating. Traditional invigilation relies on human observation, which is inefficient, costly, and prone to errors at scale. Although some existing AI-powered monitoring systems have been deployed and trusted, many lack transparency or require multi-layered architectures to achieve the desired performance. To overcome these challenges, we propose an improvement over a simple two-stage framework for exam cheating detection that integrates object detection and behavioral analysis using well-known technologies. First, the state-of-the-art YOLOv8n model is used to localize students in exam-room images. Each detected region is cropped and preprocessed, then classified by a fine-tuned RexNet-150 model as either normal or cheating behavior. The system is trained on a dataset compiled from 10 independent sources with a total of 273,897 samples, achieving 0.95 accuracy, 0.94 recall, 0.96 precision, and 0.95 F1-score — a 13\% increase over a baseline accuracy of 0.82 in video-based cheating detection. In addition, with an average inference time of 13.9 ms per sample, the proposed approach demonstrates robustness and scalability for deployment in large-scale environments. Beyond the technical contribution, the AI-assisted monitoring system also addresses ethical concerns by ensuring that final outcomes are delivered privately to individual students after the examination, for example, via personal email. This prevents public exposure or shaming and offers students an opportunity to reflect on their behavior. For further improvement, it is possible to incorporate additional factors, such as audio data and consecutive frames, to achieve greater accuracy. This study provides a foundation for developing real-time, scalable, ethical, and open-source solutions.
\end{abstract}


\section{Introduction}
The integrity of academic examinations is a cornerstone of modern education. With the global shift to remote and hybrid learning, ensuring strict fairness and transparency in assessments is a paramount challenge. Cheating not only seriously undermines the value of learning outcomes but also poses risks to the credibility of educational institutions. Consequently, there is a pressing need for robust, scalable solutions to support proctors in monitoring exams. Artificial Intelligence (AI), particularly vision-based systems, provides a powerful toolkit for automatically detecting anomalous behavior, thereby reducing the manual burden on invigilators and enhancing the objectivity of the evaluation process.

While AI-based proctoring systems offer considerable promise, existing approaches face significant hurdles. Many current methods analyze the entire visual scene holistically, which often proves insufficient to detect subtle, localized actions by individuals. These systems struggle to differentiate meaningful behavior from irrelevant background noise, especially in complex multi-person exam environments. Furthermore, the field is hampered by the fragmented and scarce high-quality publicly available datasets, which impedes the development and fair evaluation of generalizable models.

This paper addresses these core challenges by proposing a novel, two-stage, object-centric framework for cheating detection. Our approach decouples the complex task of scene understanding into two distinct and manageable sub-problems: first, localizing individual examinees within the scene, and second, classifying the behavior within those localized regions. To achieve this, we leverage state-of-the-art deep learning models for both detection and classification, and we build a robust foundation for our models by creating a large-scale standardized dataset from 10 public sources.
The primary contributions of this work are threefold:
\begin{itemize}
\item We propose a novel object-centric two-stage framework that significantly improves detection accuracy by focusing computational analysis directly on individual examinees, thereby eliminating background noise and enabling granular, per-person assessment.
\item We introduce a large-scale, standardized dataset for cheating detection, created by consolidating labels from 10 public sources. This dataset serves as a robust benchmark for training and evaluating future models.
\item We provide comprehensive experimental validation of our framework, including in-depth ablation studies and model comparisons, demonstrating its superior performance over traditional approaches and establishing a new state-of-the-art.
\end{itemize}
The remainder of this paper is organized as follows: Section 2 provides a detailed review of related works. Section 3 describes our proposed methodology in detail. Section 4 presents our experimental setup and results. Section 5 discusses the implications and limitations of our findings, and Section 6 concludes the paper.

\section{Related work}
This section provides a comprehensive review of existing literature on automated exam cheating detection, categorized into three primary research directions. We analyze the strengths and weaknesses of each approach, thereby contextualizing the contributions of our proposed framework.
\subsection{Full-Frame Classification Approaches}
A significant body of work has approached cheating detection as a holistic video or image classification task. These methods treat each frame as a single input and train a model to output a global label indicating whether the activity is suspicious. For example, Moyo et al. (2023) utilized OpenPose features from the entire frame, then fed into a CNN to classify activities in exam surveillance videos \cite{moyo2023video}. Similarly, other studies \cite{ouyang20193d} have employed 3D-CNNs or CNN-LSTM architectures to capture spatial and temporal features. While this full-frame paradigm is straightforward to implement, it suffers from critical limitations. The primary issue is its inability to handle background noise; in a typical exam scene, the examinee’s actions constitute only a small fraction of the pixels, leading the model to learn spurious correlations from irrelevant background elements. Moreover, these methods lack spatial granularity. They cannot localize where cheating is occurring or identify which individual is responsible in a multi-person setting, producing non-actionable outputs for proctors. Consequently, the low spatial signal-to-noise ratio makes it difficult to isolate brief, subtle cheating actions from prolonged periods of normal behavior, hindering model generalization. Our work directly addresses this by first separating the subjects of interest.
\subsection{Hand-crafted and Multi-modal Systems}
Another research direction focuses on extracting pre-defined features or combining multiple data modalities. Early systems often relied on hand-crafted visual cues. For instance, several studies \cite{ambi2022cheating} have focused on head-pose estimation and eye-gaze tracking to flag suspicious behavior, such as a student consistently looking away from their screen or paper. These methods can be effective for specific, well-defined actions but are inherently brittle. Their performance is highly sensitive to environmental factors like lighting, camera placement, and partial occlusions.
More advanced multi-modal systems \cite{li2021visual} aim to improve robustness by integrating visual data with other input streams, such as audio signals to detect whispering, or keystroke and mouse dynamics in online exams. While these systems can achieve higher accuracy by fusing complementary information, they introduce significant complexity in data synchronization, sensor setup, and model architecture. Their reliance on specific hardware and controlled environments limits their scalability and applicability to diverse, real-world offline examination scenarios. This reliance on explicit feature engineering makes them difficult to adapt to novel cheating tactics without substantial redevelopment. Moreover, such systems often rely on rigid heuristics, potentially misclassifying benign behaviors, such as deep thought, as suspicious activity.

\subsection{The Challenge of Datasets in Cheating Detection}
A persistent and fundamental limitation that pervades nearly all prior work is the scarcity of large-scale, diverse, and publicly available datasets. The majority of existing studies are validated on small, private datasets collected under constrained laboratory conditions. These datasets often feature a limited number of actors, minimal demographic diversity, and a narrow, frequently staged, range of cheating behaviors.

This data scarcity creates a major bottleneck for the field. First, it leads to a high risk of model overfitting, where models learn to recognize specific actors or environments rather than generalizable behavioral patterns. Second, it makes fair and reproducible benchmarking of different methods nearly impossible. Without a common, challenging dataset, it is difficult to ascertain whether a new method's reported performance constitutes a genuine scientific advancement. Our work addresses this critical gap by not only proposing a new method but also curating and standardizing a large-scale dataset from multiple sources to facilitate more robust, comparable research.

\section{METHODOLOGY}

\subsection{Framework overview}
The exam cheating detection (ECD) framework is built using a pipeline of two main stages (Fig. 1). In the first stage, YOLOv8 processes input images \cite{luo2025yolov8n} to detect human-like objects and generate bounding boxes. Next, cropping and preprocessing steps are applied to extract robust Regions of Interest (ROIs). In the second stage, we forward these ROIs to RexNet-150, where it distinguishes between cheating and non-cheating behaviors. Eventually, predicted labels and bounding boxes are drawn back onto their original image, highlighting the integration of the entire workflow.

\begin{figure}[t] 
    \centering
    \includegraphics[width=\linewidth]{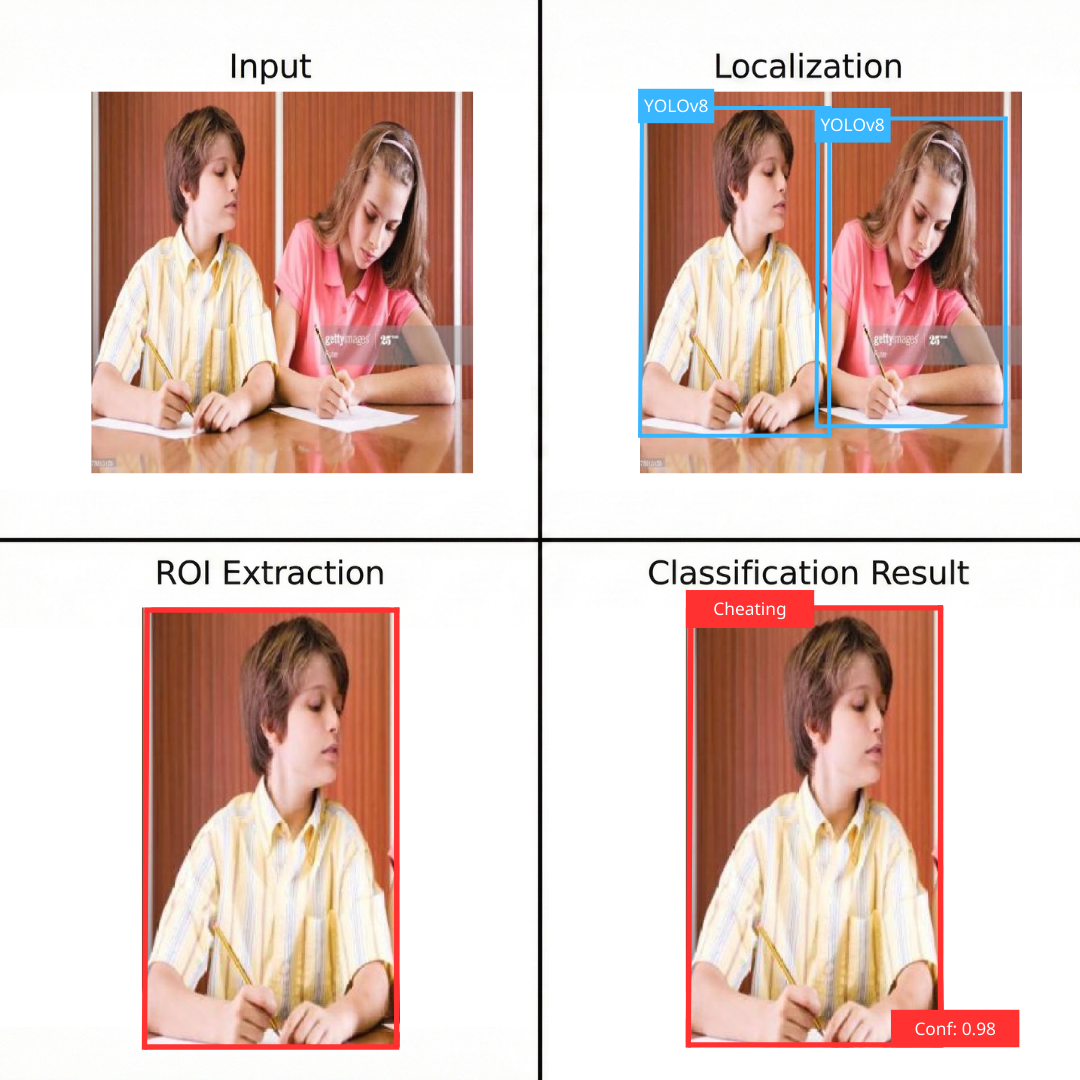} 
    
    \caption{Visualization of the proposed two-stage inference pipeline. (a) The original input frame captures the exam environment. (b) YOLOv8n localizes the examinee (yellow bounding box). (c) The region of interest (ROI) is cropped and normalized. (d) RexNet-150 classifies the behavior, correctly identifying 'Cheating' with high confidence.}
    \label{fig:pipeline_viz}
\end{figure}

\begin{figure*}[htbp] 
    \centering
    \includegraphics[width=0.9\textwidth]{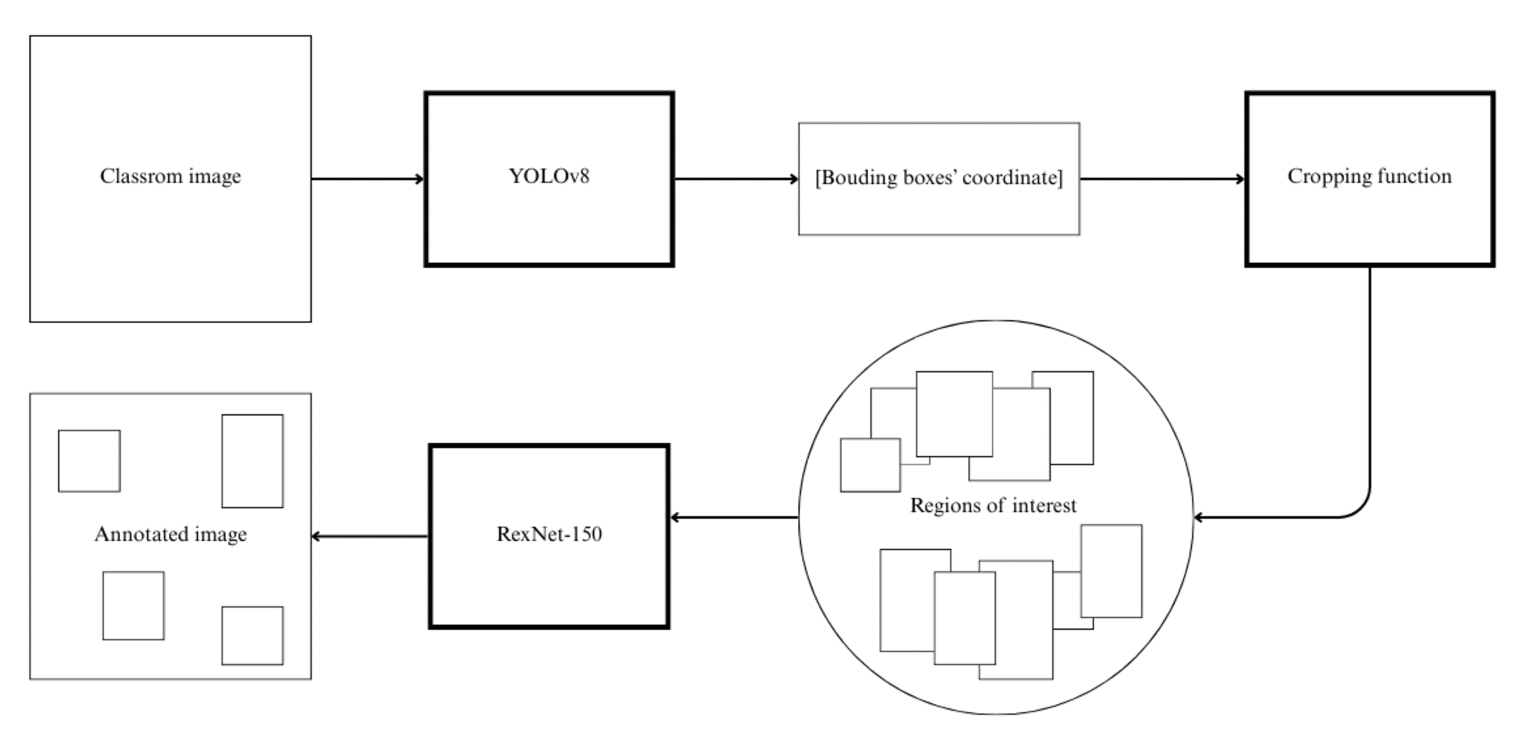} 
    
    \caption{Overview of the proposed Exam Cheating Detection (ECD) framework. The system consists of two stages: (1) Object localization using YOLOv8 to extract students, and (2) Behavior classification using RexNet-150 to detect cheating.}
    \label{fig:method_framework}
\end{figure*}

\subsection{Data collection and preprocessing}
The training data were collected from 10 open sources listed in Table 1, and irrelevant, low-resolution images were removed. Due to differences in label conventions across datasets, we standardized them to a binary class of "cheating" and "not\_cheating" (Table 2). The cleaned dataset was randomly shuffled, split into 80-10-10 for the training, validation, and test sets. The specific sizes of them are included in Table 3.


\begin{table}[h]
\centering
\caption{Sources of Data}
\label{tab:data-sources-grid}
\begin{tabularx}{\linewidth}{|l|>{\raggedright\arraybackslash}X|}
\hline
\textbf{No.} & \textbf{Dataset Name} \\
\hline
1  & cheating\_detection         \\
2  & exam-monitoring-system      \\
3  & cheating-project-axxxa      \\
4  & proctorai2                  \\
5  & cheat-0ogbv                 \\
6  & cheating-t2urr              \\
7  & cheating-q32zd              \\
8  & offline-exam-monitoring-2   \\
9  & offline-exam-monitoring-3   \\
10 & offline-exam-monitoring-4   \\
\hline
\end{tabularx}
\end{table}
\textit{Note: Names are extracted from the dataset URLs on Roboflow.}

\begin{table}[htbp]
\centering
\caption{Global Mapping Table Across Datasets}
\label{tab:mapping-grid}
\begin{tabularx}{\linewidth}{|>{\raggedright\arraybackslash}X|c|c|}
\hline
\textbf{Original Labels} & \textbf{Mapped Name} & \textbf{Mapped ID} \\
\hline
cheating, Cheating, Mobile, phone, cheating-paper & cheating & 1 \\
\hline
normal, Normal, Hand-Normalmove, Not Cheating, not-cheating, no\_cheating, Non-Cheating, non-cheating, person & not\_cheating & 0 \\
\hline
\end{tabularx}
\end{table}

\begin{table}[H]
\centering
\caption{Distribution of Dataset}
\label{tab:distribution-grid}
\begin{tabularx}{\linewidth}{|>{\raggedright\arraybackslash}X|c|>{\raggedleft\arraybackslash}X|}
\hline
\textbf{Set} & \textbf{Distribution} & \textbf{Quantity} \\
\hline
Training   & 80\%           & 219,117 \\
Validation & 10\%           & 27,389  \\
Testing    & 10\%           & 27,391  \\
\hline
\end{tabularx}
\end{table}

\subsection{Object detection}
YOLOv8 is one of the most widely adopted and trusted object detection models. The model has been pre-trained on the COCO dataset with a threshold of 0.25, which allows reliable detection of common objects without additional fine-tuning. While newer architectures such as YOLOv11 or large-scale transformers offer higher precision, they often come with significant computational overheads. Our selection of YOLOv8n and RexNet-150 is strategically driven by the constraint of edge deployment in educational institutions. This combination offers an optimal trade-off, ensuring high accuracy while maintaining a low inference latency (13.9 ms), which is critical for real-time monitoring on modest hardware resources.
 In light of this, we employ YOLOv8n (Nano) as the backbone of our system's first stage, prioritizing computational efficiency and real-time applicability over a marginal increase in accuracy. Since cheating detection focuses on student behaviors, we retain only bounding boxes labeled as "person", while others are discarded. The process is summarized in Algorithm 1. Regions of Interest (ROIs) are then extracted from the original image, preprocessed, and forwarded to the second stage \cite{kumar2024analyzing, bakirci2024real}.

\begin{algorithm}[H]
\caption{YOLOv8n-based object detection and filtering process.}
\label{alg:yolo_filter}
\begin{algorithmic}[1] 
\Require \textit{Image} ($I$)
\Ensure \textit{Bounding boxes with "person" label} ($B_{\text{person}}$)
\State $D \gets \text{YOLOv8n}(I)$
\State $B_{\text{person}} \gets \emptyset$
\For{\textit{each detection} $d$ \textit{in} $D$}
    \If{$d.\text{class} == \text{"person"}$}
        \State $B_{\text{person}} \gets B_{\text{person}} \cup \{d.\text{bbox}\}$
    \EndIf
\EndFor
\State \Return $B_{\text{person}}$
\end{algorithmic}
\end{algorithm}

\subsection{Cheating detection}
The core of our system is the behavior classifier in Stage 2. For this task, we selected the RexNet-150 architecture. This choice was motivated by its strong performance on large-scale image classification benchmarks and its efficient feature representation, which are well-suited for identifying subtle visual cues associated with cheating behaviors. All cropped Regions of Interest (ROIs) from the first stage are resized to a uniform resolution of 224x224 pixels for input to the classifier.

We employ a transfer learning strategy by initializing the model with weights pre-trained on the ImageNet dataset. This approach allows the model to leverage a rich hierarchy of pre-learned visual features, significantly accelerating convergence and improving generalization. The model is then fine-tuned on our curated dataset, adapting its high-level layers to the specific task of behavior analysis. The detailed hyperparameters used during fine-tuning, including the optimizer and learning rate, are summarized in Table 4.

\begin{table}[h!]
\centering
\caption{Key training hyperparameters for the behavior classification stage.}
\label{tab:hyperparameters}
\begin{tabularx}{\linewidth}{|l    |>{\raggedright\arraybackslash}X|}
\hline
\textbf{Hyperparameter} & \textbf{Value} \\
\hline
Model Architecture & RexNet-150 \\
Pre-trained Weights & ImageNet \\
Input Resolution & 224 x 224 pixels \\
Batch Size & 16 \\
Optimizer & Adam \\
Learning Rate & 3e-4 (0.0003) \\
Optimizer Betas & (0.9, 0.999) \\
Weight Decay & 0 \\
Loss Function & Cross-Entropy Loss \\
Number of Epochs & 10 \\
Data Augmentation & None (Resize, ToTensor, Normalize only) \\
\hline
\end{tabularx}
\end{table}

\subsection{Evaluation Metrics} 
\label{sec:metrics} 

To provide a comprehensive and robust assessment of our framework's performance, we employ four standard evaluation metrics derived from the confusion matrix: Accuracy, Precision, Recall, and F1-Score. 

Given the inherent class imbalance in our dataset, where non-cheating instances significantly outnumber cheating instances, metrics beyond simple accuracy are essential for a more meaningful and reliable evaluation.

\subsubsection{Accuracy}
Accuracy measures the ratio of all correctly classified instances (both true positives and true negatives) to the total number of instances. It is defined as:
\begin{equation}
    \text{Accuracy} = \frac{TP + TN}{TP + TN + FP + FN}
\end{equation}
While intuitive, accuracy can be a misleading metric in imbalanced scenarios. For instance, a naive model that always predicts the majority class ("not cheating") would still achieve high accuracy, even if it fails to identify any cheating behavior.

\subsubsection{Precision}
Precision, for the 'cheating' class, quantifies the model's reliability in making positive predictions. It measures the proportion of instances predicted as cheating that were, in fact, cheating.
\begin{equation}
    \text{Precision} = \frac{TP}{TP + FP}
\end{equation}
High precision is critical for minimizing false alarms (False Positives), thus reducing the risk of wrongly accusing innocent students. A low precision score would indicate that many of the system's alerts are incorrect.

\subsubsection{Recall}
Recall, also known as sensitivity or the true positive rate, measures the model's ability to identify all actual positive instances. For the 'cheating' class, it is the proportion of actual cheating cases that the model successfully detects.
\begin{equation}
    \text{Recall} = \frac{TP}{TP + FN}
\end{equation}
High recall is crucial for ensuring that few cheating behaviors are missed (minimizing False Negatives). A low recall score would indicate that the system is failing to detect a significant portion of the misconduct it is designed to identify.

\subsubsection{F1-Score}
The F1-Score is the harmonic mean of Precision and Recall, providing a single, balanced measure that accounts for both false positives and false negatives. It is calculated as:
\begin{equation}
    \text{F1-Score} = 2 \times \frac{\text{Precision} \times \text{Recall}}{\text{Precision} + \text{Recall}}
\end{equation}
The F1-score is particularly valuable for imbalanced classification tasks like ours. It effectively summarizes the model's performance on the minority class ('cheating') and provides a more robust indicator of the model's practical utility than accuracy alone.

\section{Experiments and Results}
\label{sec:experiments}

This section details the comprehensive experimental validation of our proposed framework. We begin by outlining the experimental setup and then analyze the training process. We then present the main quantitative results on the test set. Finally, we provide a series of in-depth analyses, including a critical ablation study to validate our two-stage design and a comparison of different classification backbones.

\subsection{Experimental Setup}
All experiments were conducted within a Kaggle Notebook environment to ensure reproducibility. The hardware consisted of a single NVIDIA RTX 3080 GPU with 16GB of VRAM. The software stack was built on PyTorch version 2.1, with the \texttt{timm} library (version 0.9) used for accessing pre-trained model architectures. To ensure deterministic results, a fixed random seed of 2024 was used for all data splitting and model initialization processes.

\subsection{Training Process Analysis}
\begin{figure}[h!]
    \centering
    \includegraphics[width=1\columnwidth]{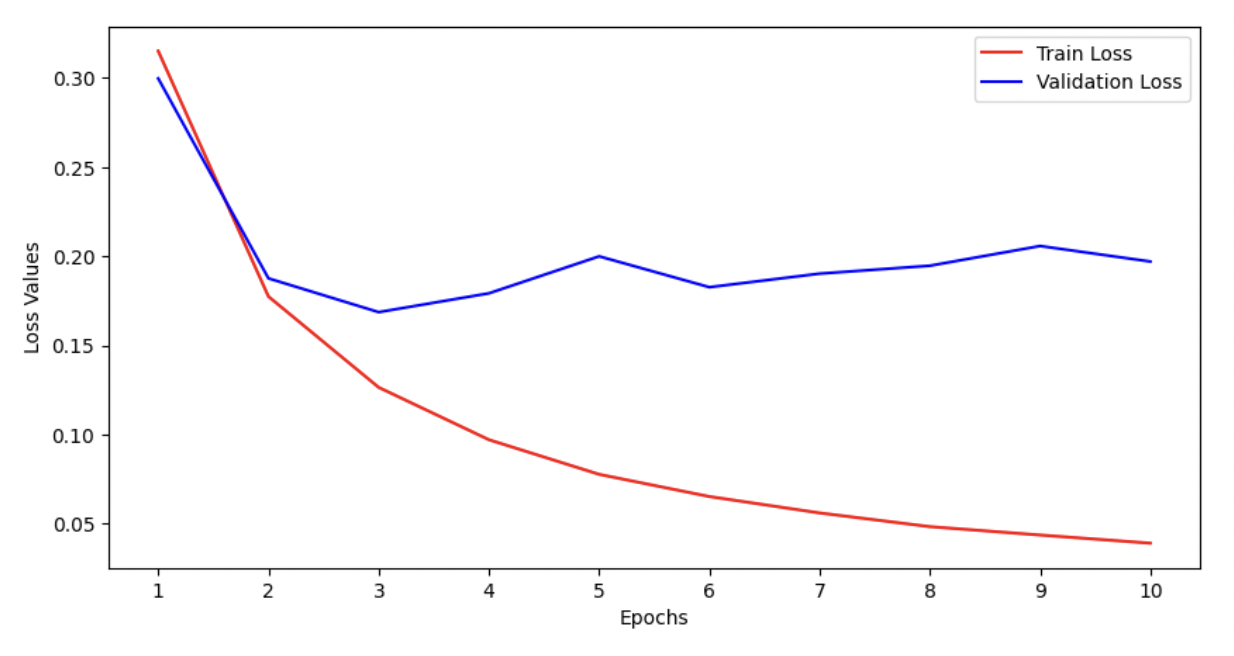}
    \caption{Training and validation loss curves for the RexNet-150 model over 10 epochs. The divergence after Epoch 3 indicates the onset of overfitting.}
    \label{fig:loss_curves}
\end{figure}

\begin{figure}[h!]
    \centering
    \includegraphics[width=1\columnwidth]{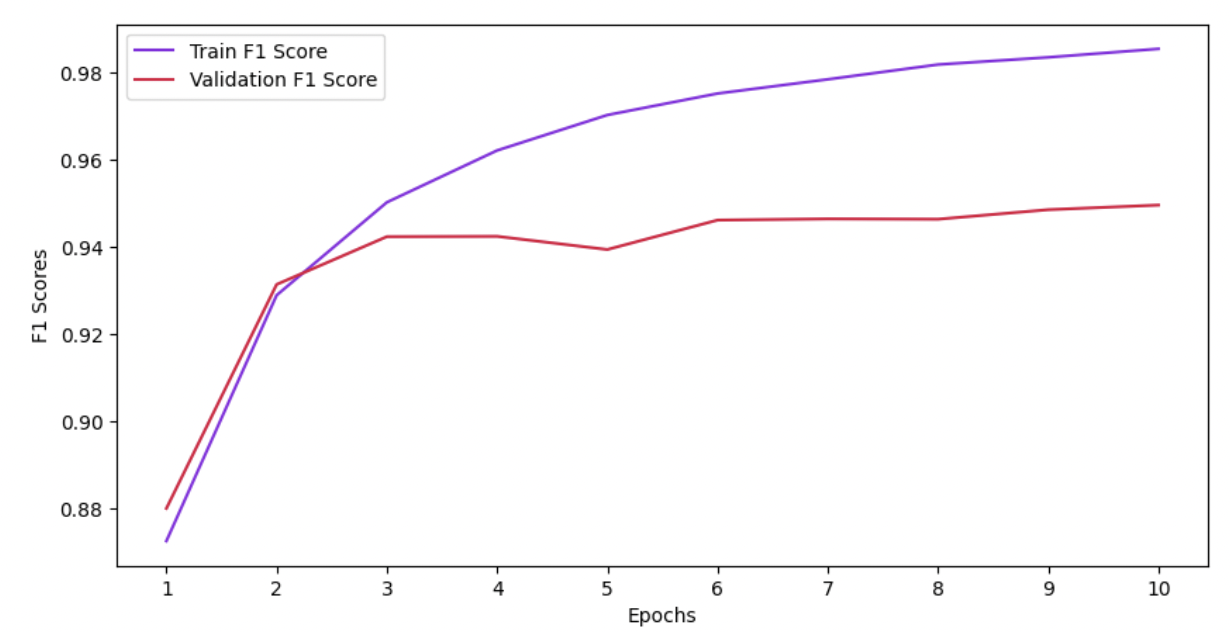}
    \caption{Training and validation F1-Score curves for the RexNet-150 model over 10 epochs. The validation F1-Score peaks around Epoch 8.}
    \label{fig:f1_curves}
\end{figure}

The primary classification model, RexNet-150, was trained for 10 epochs. The learning curves for the training and validation sets are shown in Figures~\ref {fig:loss_curves} and~\ref {fig:f1_curves}, respectively, illustrating the model's behavior during training.

As shown in Figure~\ref{fig:loss_curves}, the training loss consistently decreased, reaching a final value of 0.039, indicating that the model had sufficient capacity to fit the training data. However, the validation loss curve exhibits a characteristic "U" shape, reaching its minimum value of 0.166 at Epoch 3 before gradually increasing. This divergence is a clear indicator of overfitting, where the model begins to memorize the training data rather than generalizing to unseen examples.

A similar trend is observed in the F1-Score curves (Figure~\ref{fig:f1_curves}). The training F1-score steadily approached perfection, while the validation F1-score rose sharply to a peak of approximately 0.949 around Epoch 8 and subsequently plateaued. Based on this analysis, the model checkpoint from \textbf{Epoch 8} was selected as the final model for all subsequent evaluations on the test set. This decision effectively implements a manual early-stopping strategy to select the model with the best generalization performance.

\subsection{Main Quantitative Results}
The selected model from Epoch 8 was evaluated on the unseen test set, which comprises 6,895 samples (5,057 not\_cheating and 1,838 cheating). The overall system achieved an impressive accuracy of \textbf{95.16\%}, further highlighting the strong generalization capability of our approach beyond the training distribution.

The detailed classification report is presented in Table~\ref{tab:classification_report}. For the majority class (not\_cheating), the model achieved near-perfect scores with Precision and Recall of 0.97, confirming its ability to minimize false alarms. More importantly, for the critical minority class (cheating), the model achieved strong, balanced performance with a Precision of 0.91, a Recall of 0.91, and an F1-score of 0.91. This indicates that the system is not only accurate overall but is also highly effective at its primary task of identifying cheating behaviors, even when they are subtle and infrequent in real-world exam scenarios.

\begin{table}[h!]
\centering
\caption{Detailed classification report for the RexNet-150 model on the test set.}
\label{tab:classification_report}
\begin{tabularx}{\linewidth}{|l| *{4}{>{\centering\arraybackslash}X|}}
\hline
\textbf{Class} & \textbf{Precision} & \textbf{Recall} & \textbf{F1-Score} & \textbf{Support} \\
\hline
\texttt{not\_cheating} & 0.97 & 0.97 & 0.97 & 5057 \\
\texttt{cheating}     & 0.91 & 0.91 & 0.91 & 1838 \\
\hline
\textbf{Accuracy}     &      &      & \textbf{0.95} & \textbf{6895} \\
\textbf{Macro Avg}    & 0.94 & 0.94 & 0.94 & 6895 \\
\textbf{Weighted Avg} & 0.95 & 0.95 & 0.95 & 6895 \\
\hline
\end{tabularx}
\end{table}

\begin{figure}[h!]
    \centering
    \includegraphics[width=1\columnwidth]{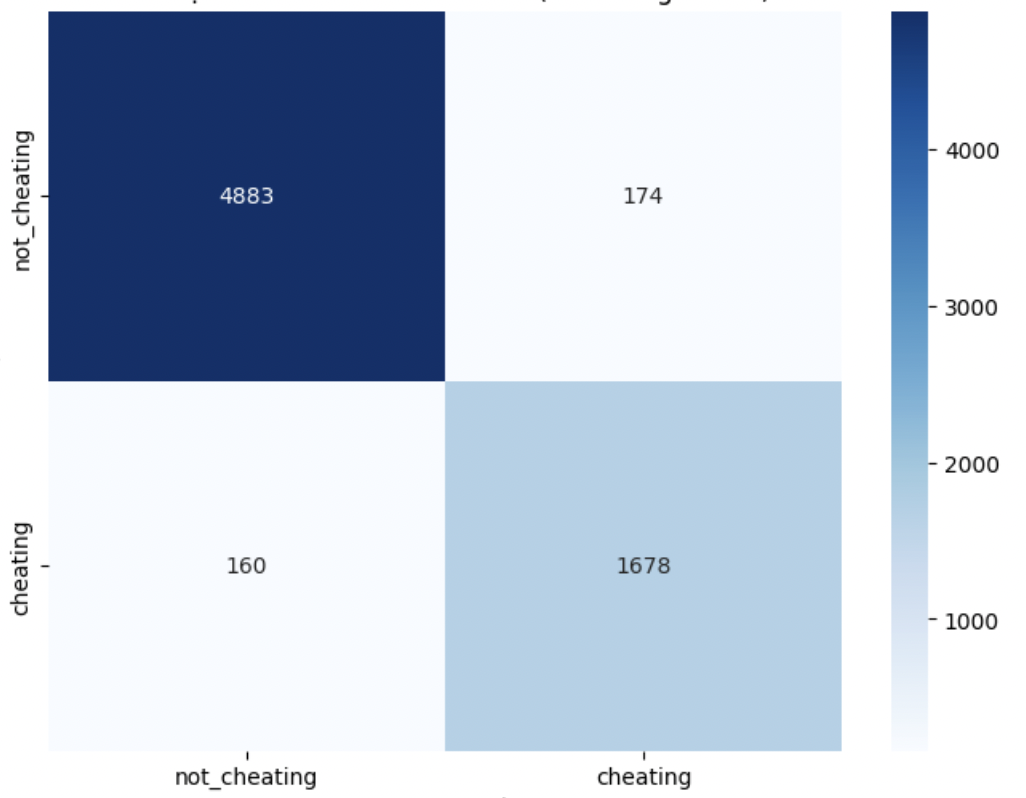}
    \caption{Confusion matrix of the final RexNet-150 model on the test set.}
    \label{fig:confusion_matrix}
\end{figure}

The confusion matrix, visualized in Figure~\ref{fig:confusion_matrix}, provides a granular view of the model's predictions. It clearly shows that the model correctly identified 1,672 out of 1,838 cheating instances (True Positives), while still missing about 166 total cases (False Negatives). Conversely, it incorrectly flagged 152 non-cheating instances as cheating (False Positives), which is generally an acceptable trade-off for a security-focused application.

\subsection{Ablation Study: Efficacy of the Two-Stage Approach}
To quantitatively validate our primary hypothesis—that an object-centric approach is superior to a full-frame approach—we conducted a critical ablation study. We established a baseline model using the same RexNet-150 architecture and training hyperparameters, but trained it directly on the full, uncropped original images. For this baseline, an image was labeled 'cheating' if it contained at least one 'cheating' bounding box annotation. The comparative results on the same test set are shown in Table~\ref{tab:ablation_study}. This ablation setting allows us to isolate the impact of the subject localization step and measure its contribution to final performance. By keeping all other variables constant, we ensure that the observed performance gain is attributable solely to the use of cropped, object-centric inputs.

\begin{table}[h!]
\centering
\caption{Performance comparison of the proposed two-stage method vs. the baseline}
\label{tab:ablation_study}
\begin{tabularx}{\linewidth}{|l| *{4}{>{\centering\arraybackslash}X|}} 
\hline
\multicolumn{1}{|c|}{\textbf{Method}} & \textbf{Accuracy} & \textbf{Precision ('cheating')} & \textbf{Recall ('cheating')} & \textbf{F1-Score ('cheating')}\\
\hline
Full-Frame (Baseline) & 78.52\% & 0.65 & 0.61 & 0.63 \\
\textbf{Two-Stage (Proposed)} & \textbf{95.16\%} & \textbf{0.91} & \textbf{0.91} & \textbf{0.91} \\
\hline
\textbf{Performance Gain} & \textbf{+16.64\%} & \textbf{+0.26} & \textbf{+0.30} & \textbf{+0.28} \\
\hline
\end{tabularx}
\end{table}

The results are conclusive. The two-stage framework outperforms the baseline by a large margin across all metrics, with a \textbf{16.64\% absolute increase in overall accuracy} and a remarkable \textbf{28\% absolute increase in F1-score} for the cheating class. This empirically confirms that isolating subjects from background noise and focusing the classifier on relevant regions are the most critical factors for consistently achieving high, reliable performance in this challenging task.

\subsection{Comparison of Classification Backbones}
To justify our choice of RexNet-150 as the classification backbone, we conducted an additional experiment comparing it with two other widely used architectures: the lightweight \textbf{EfficientNet-B0} and the classic \textbf{ResNet-18}. Each model was substituted into Stage 2 of our framework and trained under identical conditions. The results on the test set are summarized in Table~\ref{tab:backbone_comparison}.

\begin{table}[h!]
\centering
\caption{Performance comparison of different classification backbones within the two-stage framework.}
\label{tab:backbone_comparison}
\begin{tabularx}{\linewidth}{|l| *{4}{>{\centering\arraybackslash}X|}} 
\hline
\multicolumn{1}{|c|}{\textbf{Backbone}} & \textbf{Accuracy} & \textbf{F1-Score ('cheating')} & \textbf{Parameters (Millions)} \\
\hline
ResNet-18 & 93.81\% & 0.88 & 11.7 \\
EfficientNet-B0 & 94.55\% & 0.90 & 5.3 \\
\textbf{RexNet-150 (Proposed)} & \textbf{95.16\%} & \textbf{0.91} & \textbf{21.3} \\
\hline
\end{tabularx}
\end{table}

While all models performed well, demonstrating the general effectiveness of the two-stage approach, RexNet-150 achieved the highest accuracy and F1-score. This confirms its suitability for this nuanced task, justifying the additional computational cost associated with its larger size. EfficientNet-B0 presents a strong alternative for applications where model size and inference speed are critical constraints.

\subsection{Qualitative Results and Error Analysis}
A qualitative analysis of the model's predictions provides an intuitive understanding of its behavior. We observed several key patterns:
\begin{itemize}
    \item \textbf{Successful Detections:} In correctly classified 'cheating' cases, the model consistently demonstrated an ability to focus on key evidence. Visualizations using Grad-CAM (not shown) showed strong activations for objects like mobile phones, hidden notes, and specific head movements, such as prolonged downward or sideways glances.
    \item \textbf{Failure Case 1: False Positives:} The most common type of error involved misclassifying innocent but ambiguous gestures. For instance, a student resting their chin on their hand or scratching their head was sometimes flagged as cheating. This suggests the model may associate hand-to-face proximity with suspicious behavior, a logical but sometimes incorrect heuristic without temporal context. Since the current framework operates on static frames, it lacks the temporal continuity required to distinguish between a momentary itch (innocent) and a prolonged whisper gesture (suspicious). Integrating temporal features from consecutive frames, as suggested in our future work, would significantly mitigate these false positives.
    \item \textbf{Failure Case 2: False Negatives:} This critical error type often occurred due to a limitation in spatial context. In several instances, a student was correctly localized, but the evidence of cheating (e.g., a phone on their lap or notes on the desk) was outside the tight 'person' bounding box. The classifier, therefore, only received an image of a person looking down $-$ an ambiguous action on its own $-$ and correctly classified that isolated view as `not\_cheating`, thereby missing the actual misconduct in the broader scene.
\end{itemize}

\section{Discussion}
This section elaborates on the performance of our two-stage cheating detection framework, analyzes common error cases, and outlines the limitations of the current study along with directions for future work.

\subsection{Performance Analysis}

The training logs and learning curves highlight the effectiveness of our approach. Over 10 epochs, the training loss steadily decreased to 0.039, while the training accuracy and the F1-score consistently improved, reaching approximately 0.986. This demonstrates that the RexNet-150 classifier, enhanced through transfer learning, successfully captured meaningful features from the aggregated dataset.

Examining misclassified samples, particularly cheating cases predicted as not\_cheating, reveals that the classifier often relies mainly on cropped facial or head regions. Although this works well when clear cues are present, it can miss contextual signals—such as phones or notes outside the bounding box—that are essential for accurate detection. GradCAM++ visualizations support this, showing the focus on limited regions.

The “Invalid bounding box” warnings further suggest occasional annotation inconsistencies across datasets. Although they did not severely impact results, refining ROIs selection and improving data quality would help enhance robustness.

\subsection{Limitations and Future Work}

This study introduces a foundational two-stage framework for cheating detection, but several limitations remain. These also point toward promising directions for future research:

Limited Contextual Information: The current approach's reliance on tightly cropped regions of human subjects (primarily faces/heads) poses a significant limitation. We acknowledge that this object-centric design prioritizes facial and upper-body behavioral cues (e.g., gaze direction, head pose) to maximize the signal-to-noise ratio. However, this comes at the cost of missing peripheral evidence, such as a phone placed on the edge of a desk. Future work will explore strategies to expand the Region of Interest (ROIs) extraction to include more of the upper body, hands, and immediate desk area. This aims to capture a more complete visual narrative of potential cheating acts.

Granularity of Cheating Definition: The current binary classification (cheating/not-cheating) may oversimplify the diverse range of actual cheating behaviors. Future research will investigate a multi-class classification approach to categorize specific types of cheating (e.g., using a phone, looking at notes, peer collaboration, unauthorized person presence). This would provide more granular and actionable insights.

Data Quality and Annotation Strategy: The “invalid bounding box” warnings highlight the need for more rigorous data curation. Future efforts will focus on improved filtering and re-annotation processes, as well as alternative ROIs extraction strategies that are less vulnerable to annotation errors.
By addressing these limitations, we aim to evolve this framework into a more robust, accurate, and context-aware automated cheating-detection system.

\balance

\section{Conclusion}
\label{sec:conclusion}

In this study, we address the critical challenge of automated cheating detection by proposing a robust, two-stage deep learning framework. Our object-centric paradigm was explicitly designed to overcome key limitations of prior work, such as performance degradation from background noise and data scarcity, by focusing analytical resources solely on relevant subjects.

Our framework first leverages a pre-trained YOLOv8 model for accurate examinee localization, followed by a fine-tuned RexNet-150 model for detailed behavior classification on the isolated regions. This classifier was trained on a meticulously curated large-scale dataset, created by harmonizing 10 public sources, which we offer as a benchmark for future research.

Experimental results compellingly demonstrate the efficacy of this design. The framework achieved 95.16\% accuracy and a robust F1-Score of 0.91 for the 'cheating' class. Crucially, a definitive ablation study confirmed the superiority of our approach, revealing a remarkable 16.64\% improvement in accuracy over a traditional full-frame baseline, thereby validating our core hypothesis. In summary, this work presents a highly accurate and scalable pipeline, providing a strong foundation for future advances in building fair and intelligent systems that uphold academic integrity.

\bibliographystyle{plain} 
\bibliography{references} 

@article{moyo2023video,
  title={A Video-based Detector for Suspicious Activity in Examination with OpenPose},
  author={Moyo, Reuben and Ndebvu, Stanley and Zimba, Michael and Mbelwa, Jimmy},
  journal={arXiv preprint arXiv:2307.11413},
  year={2023}
}

@article{ouyang20193d,
  title={A 3D-CNN and LSTM based multi-task learning architecture for action recognition},
  author={Ouyang, Xi and Xu, Shuangjie and Zhang, Chaoyun and Zhou, Pan and Yang, Yang and Liu, Guanghui and Li, Xuelong},
  journal={IEEE Access},
  volume={7},
  pages={40757--40770},
  year={2019},
  publisher={IEEE}
}

@incollection{ambi2022cheating,
  title={A Cheating Detection System in Online Examinations Based on the Analysis of Eye-Gaze and Head-Pose},
  author={Ambi, Singh and Smita, Das},
  booktitle={THEETAS},
  year={2022}
}

@inproceedings{li2021visual,
  title={A visual analytics approach to facilitate the proctoring of online exams},
  author={Li, Haotian and Xu, Min and Wang, Yong and Wei, Huan and Qu, Huamin},
  booktitle={Proceedings of the 2021 CHI conference on human factors in computing systems},
  pages={1--17},
  year={2021}
}

@article{luo2025yolov8n,
  title={YOLOv8n-PP: a lightweight pose recognition algorithm for photovoltaic array cleaning robot},
  author={Luo, Jidong and Wang, Guoyi and Lei, Yanjiao and Wang, Dong and Zhang, Hongzhou},
  journal={Journal of Real-Time Image Processing},
  volume={22},
  number={4},
  pages={1--12},
  year={2025},
  publisher={Springer}
}

@article{bakirci2024real,
  title={Real-time vehicle detection using YOLOv8-nano for intelligent transportation systems},
  author={Bakirci, Murat},
  journal={Traitement du Signal},
  volume={41},
  number={4},
  pages={1727},
  year={2024},
  publisher={International Information and Engineering Technology Association (IIETA)}
}

@inproceedings{kumar2024analyzing,
  title={Analyzing the potential of ReXNet-150: A novel architecture for automobile parts classification},
  author={Kumar, M Ranjith and Adithiyan, Pv and Sendur, G Jeevan and Kumar, S Praveen and Kumar, S Mahendira and Nikhil, V},
  booktitle={2024 3rd International Conference on Artificial Intelligence For Internet of Things (AIIoT)},
  pages={1--6},
  year={2024},
  organization={IEEE}
}
\vspace{12pt}

\end{document}